# Online Learning to Estimate Warfarin Dose with Contextual Linear Bandits

Hai Xiao

## Abstract

Warfarin is one of the most commonly used oral blood anticoagulant agent in the world, the proper dose of Warfarin is difficult to establish not only because it is substantially variant among patients, but also adverse even severe consequences of taking an incorrect dose. Typical practice is to prescribe an initial dose, then doctor closely monitor patient response and adjust accordingly to the correct dosage. The three commonly used strategies for an initial dosage are the *fixed-dose* approach, the *Warfarin Clinical algorithm*, and the *Pharmacogenetic algorithm* developed by the IWPC (International Warfarin Pharmacogenetics Consortium). It is always best to prescribe correct initial dosage, motivated by this challenge, this work explores the performance of multi-armed bandit algorithms to best predict the correct dosage of Warfarin instead of trial-and-error procedure. Real data from the Pharmacogenetics and Pharmacogenomics Knowledge Base (PharmGKB) is used, with it a series of linear bandit algorithms and variants are developed and evaluated on Warfarin dataset. All proposed algorithms outperformed the *fixed-dose* baseline algorithm, and some even matched up the *Warfarin Clinical Dosing Algorithm*. In the end, a few promising future directions are given for further exploration and development.

————————————
Computer Science Department, Stanford University, Palo Alto, CA
Correspondence to: Hai Xiao

## 1. Introduction

Estimated More than 30 million prescriptions for Warfarin in the United States in 2004, and many more today, appropriate dose of Warfarin is only more important to both medical cost and patient safety, due to a high probability of adverse effects from incorrect dose to an individual. By study, the dose varies significantly among patients, even those look very similar, so there remains all-time interest in developing improved strategies for determining the appropriate dose (Consortium, 2009). Main efforts are consistently spent on both feature engineering (find and determine the most relevant features) and algorithms improvement. But due to the sparsely available and enough useful costly medical data, online prediction algorithms from reinforcement learning become increasingly useful in this type of problem setting.

One patient dataset is publicly available from PharmGKB, which contains 5700 patients record with Warfarin treatment from 21 research groups, cross 9 countries and 4 continents; this dataset collected demographics, background, phenotypes and genotypes, etc. total 65 features (not all 65 features exist in each record); just importantly there are 5528 patients data contains the true patient-specific optimal Warfarin dose amount through the physician-guided dose adjustment process over time. Confidently algorithm developer and researcher can use this *Therapeutic Dose of Warfarin* field as Ground Truth (Oracle) to develop and evaluate a predictive algorithm.

**Problem setting**, we discretize and classify the right dosage for patient as 3 levels according to IWPC (Consortium, 2009) for algorithmic prediction, these three categories of dosing are:

- Low Dose: under 3mg/day or 21mg/week
- Medium Dose: between 3-7mg/day or 21-49mg/week
- High Dose: above 7mg/day or 49mg/week

To frame this as a reinforcement online learning problem, we assume following rewards structure:

- $-1$: incorrect dosage (level) is predicted for the patient
- $0$: correct dosage (level) is predicted for the patient

To frame this as a Multi-Armed Bandits problem, we assume the dosage level (to predict) as bandit arm (to pull). Therefore, the problem is discretized to MAB with $K = 3$.

Further, we assume this reward for an (predicted) arm $i \in [K]$ for each patient (context) depends on his/her own features $X_t$ ($|X| = d$, $\beta_i \in \mathbb{R}^d$):
$$r_t(X_t, i) = X_t^T \beta_i + \varepsilon_{i,t}$$
$\varepsilon_{i,t}$ is an independent random variable/noise term, $\beta_i$ is an unknown coefficient parameters for sample features.

Therefore, when using linear bandit, the goal of the problem is to design a bandit algorithm that learns the policy mapping:
$$\pi : X \to i$$
That yields the maximal expected rewards.

Assume optimal policy $\pi^*$ always ($\forall t$) yields the maximum expected reward across patients:
$$\pi^* : X \to a^* = \max_{a \in K}(X_t^T \beta_a)$$

With this setup, the goal is to create and evaluate an algorithm that minimize the cumulative expected regret:
$$R_T = \sum_{t=1}^{T} \mathbb{E}[\max_{a \in K}(X_t^T \beta_a) - X_t^T \beta_i]$$
$i \in K$, is the arm chosen by agent at timestamp $t$.

## 2. Related Work

In this section, previous work, approaches, and relevant methods are provided to serve the scope and background for the work in this study.

### 2.1 Reference Baseline to Warfarin Dose

Primarily there are three established algorithms used for an initial dosage, those are

1. Fixed-dose approach:
   Assign 5mg/day or 35mg/week to all patient. Benefit of doing so is that **the worst possible distance** between given dosage and a patient's true optimal dosage has a smaller bound, since the prescribed dosage is always at medium. But this approach has very strong bias due to lack of learning and agnostic to any patient features, so in term of predictive error, it is unbounded, and depend upon test population (for example when none of the patients' true dosage is between 21-49mg/week, when use discretized dosage level, an error rate would be 100%).

2. Warfarin Clinical Dosing Algorithm (**WCDA**):
   This is a simple but effective linear model based on patent's 8 features: age, height, weight, race and medications exposure, etc. Its developed by Medicine researchers in Consortium. The model is linear regression one with an output square to predict weekly dosage in mg.

3. Warfarin Pharmacogenetic Dosing Algorithm:
   Another linear model proposed by Consortium, main difference with *WCDA* is that **WPDA** also includes genotype as feature input.

Algorithmic details regarding *WCDA* and *WPDA* are located in sec. *S1f* & *S1e* of attached Supplementary Appendix Supplement to: The International Warfarin Pharmacogenetics Consortium. Estimation of the Warfarin dose with clinical and pharmacogenetic data. N Engl J Med 2009;360:753-64.

## 2.2 Relevant Methods and Approaches

In different application domains, many proposals exist for similar setting but different problems, for example Contextual-Bandit Approach had been widely adapted to personalize news article recommendation, representative lectures include [Lihong Li, Wei Chu, Robert E. Schapire, et al.] Contextual Bandits with Linear Payoff Functions[1], A Contextual-Bandit Approach to Personalized News Article Recommendation.

To non-contextual Multi-Armed bandit problem, upper confidence bound (UCB) algorithms are proved optimal [Lai and Robbins, et al. 1985] if the Rewards are i.i.d. By Keeping upper bounds on the best rewarded potential arms, then pull arm with the highest upper bound, UCB makes good tradeoff between Exploitation vs. Exploration.

But many real MAB problem settings are rather Contextual vs. pure random. Contextual bandits extend MAB by making the decision conditional on the state (or observation) of the environment. During each iteration, learning agent first access certain relevant information, called *context,* from the environment, which is used to select action. Contextual information commonly used are user feature or profile, in our problem setting this can be patient features.

Combined this with linear payoff (rewards) wrt. feature $X_{t,a}$ we have linear disjoint (arm has own parameters $\theta_a$) **LinUCB** theoretically framed as:

Reward Assumption:
$$\mathrm{E}[r_{t,a}|X_{t,a}] = X_{t,a}^T \theta_a^*$$

Parameter Estimation:
$$\hat{\theta}_a = (D_a^T D_a + I_d)^{-1} D_a^T c_a \quad \text{(Ridge Regression)}$$

Bound of the Variance:
$$|X_{t,a}^T \hat{\theta}_a - X_{t,a}^T \theta_a^*| \leq \alpha \sqrt{X_{t,a}^T (D_a^T D_a + I_d)^{-1} X_{t,a}}$$

Decision making (pick $a^*$, s.t.):
$$\underset{a}{\mathrm{argmax}}(X_{t,a}^T \hat{\theta}_a + \alpha \sqrt{X_{t,a}^T (D_a^T D_a + I_d)^{-1} X_{t,a}})$$

This gives **LinUCB** algorithm [Yisong Yue, Caltech]:

```
Algorithm 1 LinUCB with disjoint linear models.
 0: Inputs: c_t ∈ R_+
 1: for t = 1, 2, 3, ..., T do
 2:    Observe features of all arms a ∈ A_t: x_{t,a} ∈ R^d
 3:    for all a ∈ A_t do
 4:       if a is new then
 5:          A_a ← I_d (d-dimensional identity matrix)
 6:          b_a ← 0_{d×1} (d-dimensional zero vector)
 7:       end if
 8:       θ̂_a ← A_a^{-1} b_a
 9:       p_{t,a} ← θ̂_a^⊤ x_{t,a} + c_t √(x_{t,a}^⊤ A_a^{-1} x_{t,a})
10:    end for
11:    Choose arm a_t = arg max_{a ∈ A_t} p_{t,a} with ties broken arbitrarily, and observe a real-valued payoff r_t
12:    A_{a_t} ← A_{a_t} + x_{t,a_t} x_{t,a_t}^⊤
13:    b_{a_t} ← b_{a_t} + r_t x_{t,a_t}
14: end for
```

On top of this [Chou, Chiang, Lin, Lu] further explored the idea to also feed the learning agent some pseudo rewards on non-selected arms after each action. Motivated by the facts that a better performance can be normally achieved if another hypothetic rewards to (or some) the non-selected actions can be revealed to learner as well. They proposed a new framework that feeds the agent with an over-estimated pseudo-rewards on non-selected actions, and a forgetting mechanism to balance the negative influence of the introduced over-estimation in the long run, coupling these ideas with LinUCB, they designed a algorithm called linear pseudo-rewards upper confidence bound (**LinPRUCB**). Their experiments show that LinPRUCB shares the same order of regret bound to LinUCB but enjoys some faster reward gathering in the earlier iterations, which yields faster computation.

## 3. Approach Algorithms

To tackle the challenge of estimating Warfarin Dose with fast online learning, a few algorithms and approaches are studied here and proposed to experiment later.

First meaningful baselines need to be established for any upcoming algorithmic work.

### 3.1 Baseline Establishment to Warfarin Dose

To evaluate a series of algorithms, best practice is to establish a baseline upfront, along with this process, it is also important to establish nearly identical data set and features set feeding to the algorithms under study.

There are **two baselines** to be used in this work: *Fixed-dose* and *WCDA*. Post features transform and sanitization, there are 4386 total records of data left with complete set of $X$ (needed features from more strictive *WCDA vs. Fixed-dose*) and $y$ (the available Ground Truth Dosage). Therefore, along the course of remaining study, this subset of 4386 entries are used all over the places.

To *Fixed-dose* algorithm, I bucketized real dosage to 3 levels according to this scheme: $\{<21:0, 21-49:1, >49:2\}$, and algorithm simply predicts $\hat{y} = 1$ for all patients.

To *WCDA* algorithm, In addition to the same discretization of real continuous dosage, it is also necessary to transform non-scalar inputs type to numeric, and deal missing values with care. Since *WCDA* is a linear regression model built by expert, so even not a RL model, it is still an ideal baseline to benchmark online learning fast RL models to this problem.

### 3.2 LINUCB (with the same feature set as *WCDA*)

This is the 1st fast online learning algorithm implemented in this study. It follows **Algorithm 1** from previous page, and papers from [Wei Chu, Lihong Li, Lev Reyzin, and Robert E. Schapire] and [Feng Bi, Joon Sik Kim, Leiya Ma, Pengchuan Zhang].

Given the problem setting, and preliminary knowledge in medication, I do not split feature set among arms or actions, this simplification shall not diminish the value of remaining work, and it is possible to try out different feature set for different actions in the future. Provided this, all $X_{t,a} \longrightarrow X_t$ throughout this writing.

Another note is what I implemented is a dis-joint linear LINUCB model, where each action has got its own parameter Matrix, this is well described in **Algorithm 1** vs. [Wei Chu, Lihong Li, et al] paper.

### 3.3 LINPRUCB (with the same feature set as *WCDA*)

Appreciated the idea of Pseudo-Reward, over-estimate and forgetting mechanism for fast learning, an implementation of LINPRUCB is also provided and analyzed in this study. **N.B. Algorithm 2** below is sketched from paper [Chou, Chiang, Lin, Lu], but with my observed fix of additional line 9 and 13 (which was missing). Also, it needs to set proper regularize constant in line 11.

**Algorithm 2** Linear Pseudo-Reward UCB (LINPRUCB)

0: Parameters: $\alpha > 0, \beta > 0, \eta < 1$
1: Initialize: $W_{1,a} := \mathbf{0}_d, \widehat{Q}_{0,a} := \mathbf{I}_d, \widehat{V}_{0,a} := V_{0,a} := \mathbf{0}_{d \times d}$
   $\widehat{Z}_{0,a} := Z_{0,a} := \mathbf{0}_d,$ for every $a \in [K]$
2: **for** $t := 1$ **to** T **do**
3:   Observe $X_t$
4:   Select $a_t := \underset{a \in [K]}{\operatorname{argmax}} W_{t,a}^T X_t + \alpha \sqrt{X_t^T \widehat{Q}_{t-1,a}^{-1} X_t}$
5:   Receive reward $r_{t,a_t}$
6:   **for** $a \in [K]$ **do**
7:     **if** $a = a_t$ **then**
8:       $V_{t,a} := V_{t-1,a} + X_t X_t^T$
9:       $Z_{t,a} := Z_{t-1,a} + X_t r_{t,a_t}$
10:    **else**
11:      $p_{t,a} := \max[-1, \min(W_{t,a}^T X_t + \beta \sqrt{X_t^T \widehat{Q}_{t,a}^{-1} X_t}, \mathbf{0})]$
12:      $\widehat{V}_{t,a} := \eta \widehat{V}_{t-1,a} + X_t X_t^T$
13:      $\widehat{Z}_{t,a} := \eta \widehat{Z}_{t-1,a} + X_t p_{t,a}$
14:    **end if**
15:    $\widehat{Q}_{t,a} := \mathbf{I}_d + V_{t,a} + \widehat{V}_{t,a}$
16:    $W_{t+1,a} := \widehat{Q}_{t,a}^{-1} (Z_{t,a} + \widehat{Z}_{t,a})$
17:  **end for**
18: **end for**

### 3.4 Trinary vs. Binary Reward Structure

Reward Structure: reward is currently defined as binary discrete values {-1, 0}, one potential caveats of this aggressive discretization are that models will tend to ignore bigger discrepancy between prediction and true dosage, leading to potential severer consequences, e.g. predict one patient of medium true dose to either low or high dose will incur the same model regret to predict a patient of low real dosage to high, vice versa.

Between use binary rewards and use the real-valued rewards to count a predictive error in mg/week, it is interesting to check how models behave with an increased discrete rewards level.

Follow the idea, a Trinary Reward Structure is then tried with rewards {-2, -1, 0}, where reward is -2 if discrepancy between prediction and truth dose is 2 levels, -1 if just 1 level, or otherwise 0.

Expectedly, this should address some of the caveats listed above.

### 3.5 Extra Approaches and Fused Hybrid Models

There are a couple of other interesting ideas or algorithmic modifications implemented in this work, which includes:

- Explore more input features to L<small>IN</small>UCB or L<small>IN</small>PRUCB (add *Gender* to *WCDA* features)
- Combine previous algorithms or approaches, e.g. Trinary Rewards + L<small>IN</small>UCB or L<small>IN</small>PRUCB

### 3.6 Some Approaches for Future Exploration

Still couple of very interesting ideas worthy to explore in the later work:

- Continuous Rewards Structure
- Incorporate baseline in online model to decrease variance, similar idea to A3C
- Non-linear Payoff Function Bandits

### 4. Implementation Methods

This section provides most relevant details with regarding to implementations in this work.

- Computational Method
  Given the problem setting, domain dataset and algorithms to explore, Neuron Network is not needed or used (as Deep RL), neither any Deep Learning Framework.
- Software Utilities
  Since mostly linear and matrix operations, and small dataset size, Numpy, Pandas Data Frames, and Matplotlib etc. software are adequately efficient to the problem.
- Data Preprocess
  Input features to our models are digitized, it also includes null handling & one-hot encoding for categorical feature transforms whenever necessary. To make sense from model compare, use *WCDA* features around
- Data Usage Model
  Provided the dataset is small, and our online fast RL setting, dataset is not split to training, validation and test set as in traditional supervised learning problem. Instead, these models keep learning as they iterate through full dataset once. Among multiple rounds same dataset is randomly shuffled to simulate random order of input samples in real world. Model parameters accumulative updates don't persist through different rounds.
- Data Collection and Evaluation Method
  During the process of learning, build up following measurement:
  o Running Accuracy (both last N steps and till now: $t_0 \to t$)
  o Running Regret (both last N & $t_0 \to t$)
  o Measurement data cross multiple runs (samples shuffled independently), with statistic support including Confidence Interval, standard deviation, etc. for robust model evaluation.

# Online Learning to Estimate Warfarin Dose with Contextual Linear Bandits

## 5. Experiment Results

### 5.1 Baseline

First ***Fixed-dose*** and ***WCDA*** algorithm are implemented as baselines for upcoming algorithmic work. To baseline, both the running average accuracy & regret are computed, over both last Bt (100) steps and whole history ($t_0 \rightarrow t$). Diagram below shows these over one round (episode) run through all post selected samples (~4386 patient data):

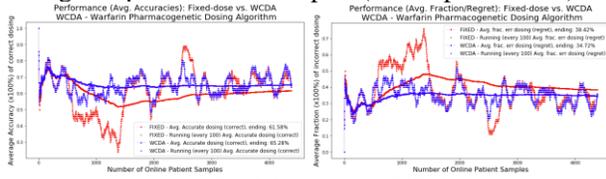

Figure 1. ***Fixed-dose*** (red) vs. ***WCDA*** (blue)
running average Accuracy (L) vs. Regret (R)

In Figure 1, smoother curves are for averages over history. Observation here is that ***WCDA*** has better avg. accuracy & lower regret than ***Fixed-dose*** from any 1 time run through.

To confirm the robustness of this observation, multiple independent rounds (episode) of running through over randomly shuffled sample set is also simulated with no state carrying over different rounds; N=100 is chosen (100 randomly shuffled episodes) so that a meaningful Confidence Interval can be interpreted. Here comes plot:

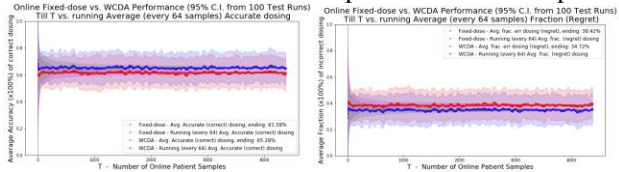

Figure 2. ***Fixed-dose*** (red) vs. ***WCDA*** (blue)
running average Accuracy (L) vs. Regret (R) with C.I.

In Figure 2 band of 95% C.I. is displayed with mean curve we clearly confirm the earlier observation that ***WCDA*** has better avg. accuracy & lower regret than ***Fixed-dose***

During this process, we come to following performance numbers:           ***WCDA*** vs. ***Fixed-dose***
*Avg. Accuracy (approx.)*      65.28%      61.58%
*Avg. Regrets (approx.)*       34.72%      38.42%

In addition, Total expected regrets are also plotted below:

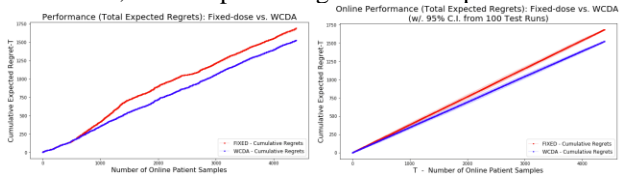

Figure 3. ***Fixed-dose*** (red) vs. ***WCDA*** (blue)
Total Regrets till Time Step t (L – 1 run) vs. (R – 100 runs with C.I.)

### 5.2 LINUCB

LINUCB is implemented using **Algorithm 1**. with a very simple tune of α. It is different with [Lihong Li, Wei Chu] paper in that disjoint actions (parameter) are implemented. Diagram below shows ***LINUCB*** vs. ***Fixed-dose*** over one round through all selected samples (~4386).

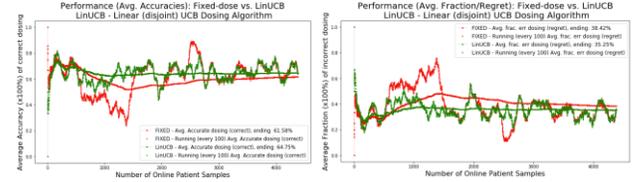

Figure 4. ***Fixed-dose*** (red) vs. ***LINUCB*** (green)
running average Accuracy (L) vs. Regret (R)

In Figure 4, smoother curves are for averages over history. Observation here is that ***LINUCB*** has better avg. accuracy & lower regret than ***Fixed-dose*** with an one time through.

Another diagram shows ***LINUCB*** vs. ***WCDA*** over one round through all selected samples (~4386).

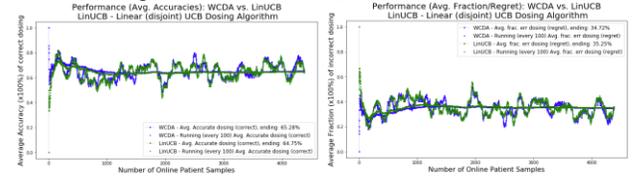

Figure 5. ***WCDA*** (blue) vs. ***LINUCB*** (green)
running average Accuracy (L) vs. Regret (R)

Figure 5 shows that ***LINUCB*** performs very close (almost same) with ***WCDA*** on both running avg. accuracy & regret

To further confirm the robustness of ***LINUCB*** algorithm and implementation, it is also run through independently random-shuffled samples 100-round with no state carry-over. Diagram below shows the comparison of prediction accuracy among ***WCDA*** (blue) vs. ***LINUCB*** (green) vs. ***Fixed-dose*** (red) with band of 95% C.I.

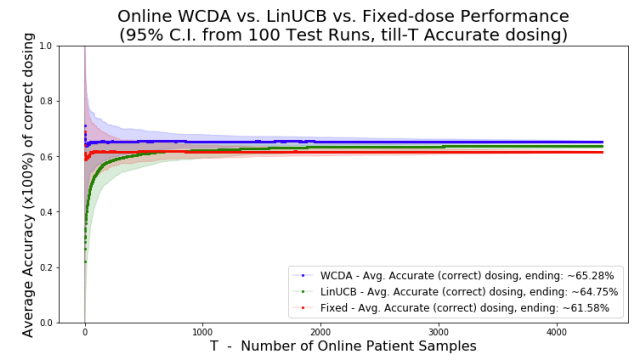

Figure 6. ***WCDA*** (blue) vs. ***LINUCB*** (green) vs. ***Fixed-dose*** (red)
running average Accuracy (Time Step 0 → t)

Another diagram below shows the comparison of running regret among *WCDA* (blue) vs. *LinUCB* (green) vs. *Fixed-dose* (red) with band of 95% C.I.

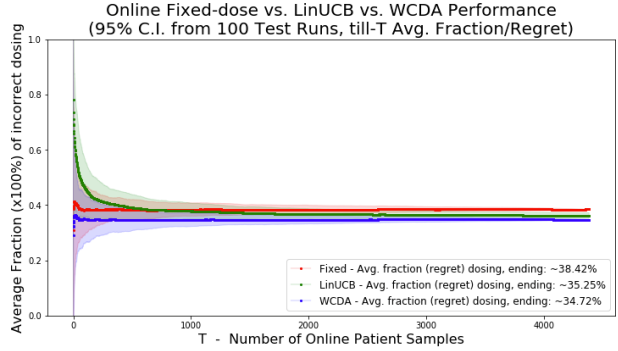

Figure 7. *WCDA* (blue) vs. *LinUCB* (green) vs. *Fixed-dose* (red) running average Regret (Time Step 0 → t)

Figure 6 and 7 show that *LinUCB* indeed performs really close to *WCDA* on both running avg. accuracy and regret; And *LinUCB* outperforms *Fixed-dose* with a good margin (> 3.0%)

Until this point, we have following performance numbers:

|  | *WCDA* | vs. *LinUCB* | vs. *Fixed-dose* |
|---|---|---|---|
| Avg. Accuracy (approx.) | 65.28% | 64.75% | 61.58% |
| Avg. Regrets (approx.) | 34.72% | 35.25% | 38.42% |

Finally, diagram plotted below shows cumulative expected Regret-T among *WCDA* vs. *LinUCB* vs. *Fixed-dose* with 95% Confidence Interval:

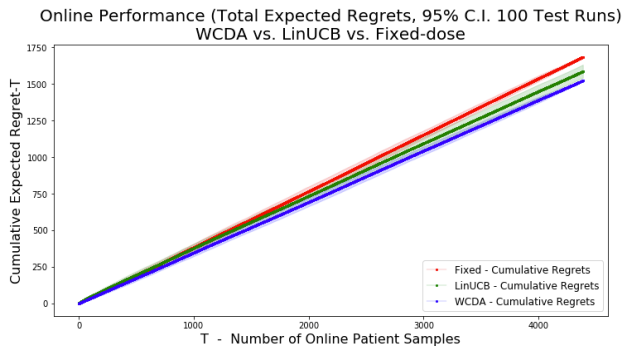

Figure 8. *Fixed-dose* (red) vs. *LinUCB* (green) vs. *WCDA* (blue) Total Regrets till Time Step t (95% C.I. cross 100 runs)

As seen here, *LinUCB* performs better than *Fixed-dose*, close to but still not better than *WCDA*.

Another observation is that cumulative regrets seem to be linear in all these 3 cases. It is a motivation to look for a better algorithm or methodology.

### 5.3 Further Directions and Exploration

#### 5.3.1 Adding More Features

So far, the same *WCDA* input features are provided to the *LinUCB* model, consider 'Gender' may be useful, its one-hot encodings (2 extra fields) are inserted to the feature $X$, this leads to a new model of *LinUCB* of 11 vs. 9 features:

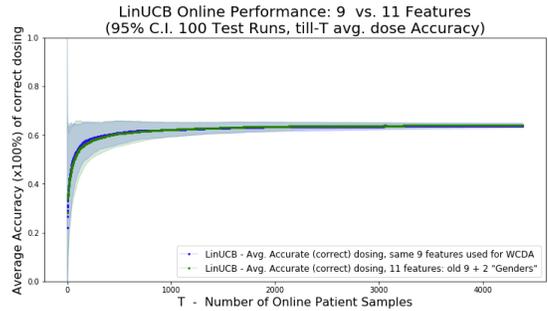

Figure 9. *LinUCB-9* (blue) vs. *LinUCB-11* (green) running avg. Accuracy (TS 0 → t, 100 runs, 95% C.I.)

Figure 9 tell us that just adding 'Gender' along seems to be a little beneficial, but the gain is very marginal.

#### 5.3.2 LinPRUCB — Adding Pseudo Rewards

Use **Algorithm 2** (added line 9/13, and modified line 11 to regularize pseudo reward $p_{t,a} \in [-1,0]$), model performed as

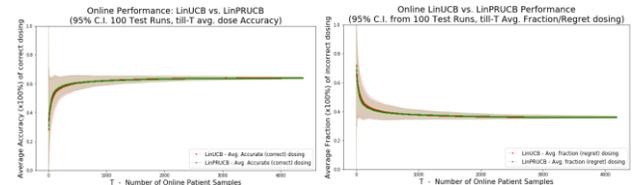

Figure 10. *LinUCB* (red) vs. *LinPRUCB* (green) running avg. Accuracy (L) vs. Regret (R) with C.I.

Figure 10 tells that *LinPRUCB* performs to higher accuracy, or lower regret slightly faster (visibly) compare to *LinUCB*, but gap is small, may due to very preliminary α, β, η tuning

#### 5.3.3 Trinary Rewards — Modify Reward Struct

Knowing the caveat of binary rewards, a $[-2, -1, 0]$ reward (yet continued) is introduced to penalize bigger error more with its implementation performance measurement below:

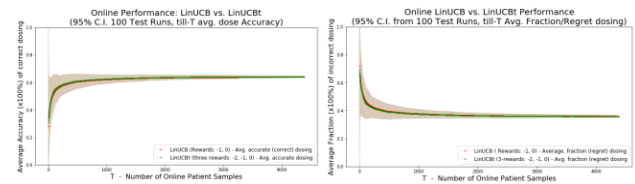

Figure 11. *LinUCB* (red) vs. *LinUCBT* (green) running avg. Accuracy (L) vs. Regret (R) with C.I.

Figure 11 shows slightly visible improvement from trinary rewards model (*LinUCBT*) over its binary counterpart (*LinUCB*)

### 5.3.4 LINPRUCBT — Trinary with Pseudo Rewards

**LINPRUCBT** is a hybrid model combining Trinary rewards and Pseudo rewards model (**Algorithm 2** line-11 regularized to $p_{t,a} \in [-2,0]$), hope to get more feedback gain from both:

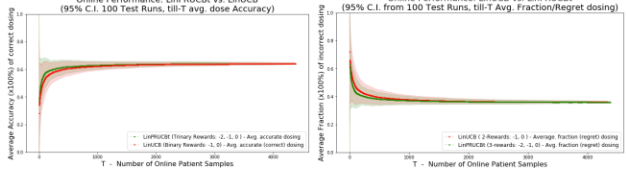

Figure 12. *LINUCB* (red) vs. *LINPRUCBT* (green) running avg. Accuracy (L) vs. Regret (R) with C.I.

Figure 12 shows notable improvement from *LINPRUCBT* compare to org. *LINUCB*, especially at earlier iterations ($t < 800$) hybrid model performs to better notably faster, this is what we usually prefer for fast online learning algorithms.

Finally, diagram plotted below shows cumulative expected Regret-T among *WCDA* vs. *LINPRUCBT* vs. *LINUCB* vs. *Fixed-dose* with 95% Confidence Interval:

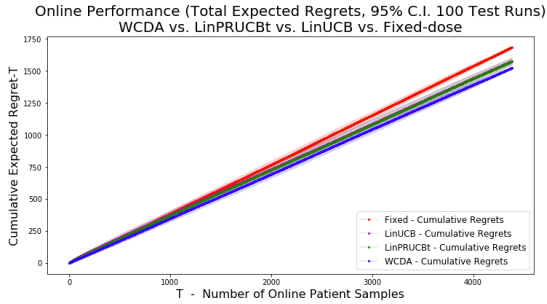

Figure 13. *Fixed-dose* (red) vs. *LINUCB* (magenta) vs. *LINPRUCBT* (green) vs. *WCDA* (blue) Total Regret till timestep t (95% C.I. 100 runs)

Show here and compare Figure 8, *LINPRUCBT* performs second to *WCDA*, better than both *LINUCB* and *Fixed-dose*

### 5.3.5 Side-by-Side Model Algorithms Comparison

Capturing improvement trends from developed algorithms, Figure 14 and 15 below provide progressive performance improvement along different online timestep with 95% C.I. These Plots have clearly showed progressive improvement of the developed algorithms and their comparing positions with two prior baselines.

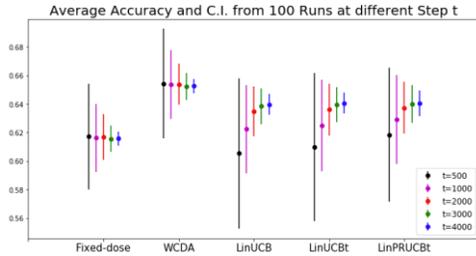

Figure 14. *Fixed-dose* vs. *WCDA* vs. *LINUCB* vs. *LINUCBT* vs. *LINPRUCBT* Running avg. Accuracy at Timestep t=500/1000/2000/3000/4000 (95% C.I.)

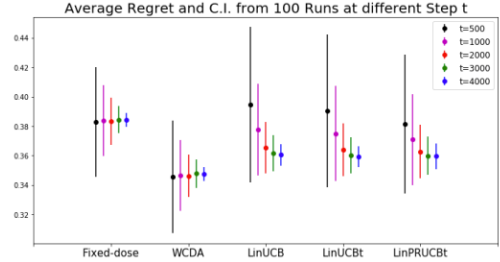

Figure 15. *Fixed-dose* vs. *WCDA* vs. *LINUCB* vs. *LINUCBT* vs. *LINPRUCBT* Running avg. Regrets at Timestep t=500/1000/2000/3000/4000 (95% C.I.)

Figure 16. below shows Total Expected Regrets at different time step by different algorithms, lower better, v-line shows C.I. range

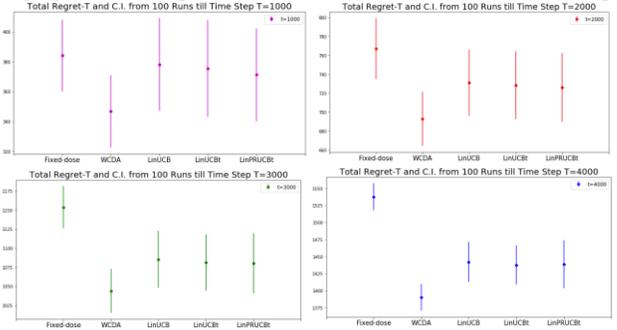

Figure 16. *Fixed-dose* vs. *WCDA* vs. *LINUCB* vs. *LINUCBT* vs. *LINPRUCBT* Total Expected Regret from timestep $0 \to t = $ 1000/2000/3000/4000 (95% C.I.)

## 6. Conclusion

Started with two established & commonly used baselines, to solve MAB with UCB and contextual linear payoff, and built on top of LINUCB & it's extensions, this work has demonstrated series of fast online learning algorithm with progressive performance improvement.

With processed, comparable feature set from PharmGKB dataset, all algorithms overbeat *Fixed-dose* baseline, and reach close enough to *WCDA*, with the hybrid algorithm **LINPRUCBT** performs the best, especially better at early iterations than **LINUCB**.

The performance gain of **LINPRUCBT** shows the benefits from both pseudo rewards and trinary vs. reward structure. With cumulative regrets still linear & directions remain to explore, it is quite possible to acquire better solution late.

## 7. Acknowledgements

I gratefully thank Stanford Assistant Professor Emma Brunskill, and research assistants Ramtin Keramati, Chelsea Sidrane, Sudarshan Seshadri, et al. for their help on the problem definition, support and valuable feedback.

# References

Hamsa Bastani, and Mohsen Bayati. Online decision-Making with High-Dimensional Covariates. 2015. Stanford University.
http://web.stanford.edu/~bayati/papers/lassoBandit.pdf

The International Warfarin Pharmacogenetics Consortium. N Eng J Med 2009. Estimation of the Warfarin Dose with Clinical and Pharmacogenetic Data.
https://www.nejm.org/doi/full/10.1056/NEJMoa0809329

Tor Lattimore and Csaba Szepesvári. Bandit Algorithms.
https://tor-lattimore.com/downloads/book/book.pdf
https://banditalgs.com

Wei Chu, Lihong Li, Lev Reyzin, and Robert E. Schapire. Contextual Bandits with Linear Payoff Functions.
http://proceedings.mlr.press/v15/chu11a/chu11a.pdf

Ku-Chun Chou, Chao-Kai Chiang, Hsuan-Tien Lin, Chi-Jen Lu. Pseudo-reward Algorithms for Contextual Bandits with Linear Payoff Functions. JMLR: Workshop and Conference Proceedings 29:1-19, 2014 ACML 2014
https://www.csie.ntu.edu.tw/~htlin/paper/doc/acml14pseudo.pdf

Lihong Li, Wei Chu, John Langford, and Robert E. Schapire. A Contextual-Bandit Approach to Personalized News Article Recommendation.
https://arxiv.org/pdf/1003.0146.pdf
https://pdfs.semanticscholar.org/dd9f/73a7701489c2d82d7fdbb326e39a93ae6a0d.pdf

Alekh Agarwal, Daniel Hsu, Satyen Kale, John Langford, Lihong Li, and Robert E. Schapire. Taming the Monster: A Fast and Simple Algorithm for Contextual Bandits.
https://arxiv.org/pdf/1402.0555v2.pdf

I.W.P.Consortium. Estimation of the warfarin dose with clinical and pharmacogenetic data. New England Journal of Medicine, 360(8):753-764, 2009.

Feng Bi, Joon Sik Kim, Leiya Ma, Pengchuan Zhang.
http://www.yisongyue.com/courses/cs159/lectures/LinUCB.pdf
Contextual Linear Bandit Problem & Applications. Caltech
http://www.yisongyue.com/courses/cs159